\pdfoutput=1

\documentclass[11pt]{article}

\usepackage{emnlp2021}

\usepackage{times}
\usepackage{latexsym}
\usepackage{pgfplots}
\pgfplotsset{width=7cm,compat=1.9}

\usepackage[T1]{fontenc}

\usepackage[utf8]{inputenc}

\usepackage{microtype}

\title{Miðeind's WMT 2021 submission}

\author{Haukur Barri Símonarson, Vésteinn Snæbjarnarson, \\ {\bf Pétur Orri Ragnarsson, Haukur Páll Jónsson and Vilhjálmur Þorsteinsson}\\
 Miðeind ehf., Reykjavík, Iceland \\
  \texttt{\{haukur, vesteinn, petur, haukurpj, vt\}@mideind.is} \\}

\begin{document}
\maketitle
\begin{abstract}
We present Miðeind's submission for the English$\to$Icelandic and Icelandic$\to$English subsets of the 2021 WMT news translation task. Transformer-base models are trained for translation on parallel data to generate backtranslations iteratively. A pretrained mBART-25 model is then adapted for translation using parallel data as well as the last backtranslation iteration. This adapted pretrained model is then used to re-generate backtranslations, and the training of the adapted model is continued.
\end{abstract}

\section{Introduction}
Our work on machine translation for Icelandic has been going on for a couple of years as a part of the government sponsored Icelandic Language Technology Programme \cite{Nikulsdttir2020LanguageTP}. By building on state-of-the-art solutions we have developed an open and effective translation system between Icelandic and English.

To achieve this, we collect parallel Icelandic and English texts which are filtered for good quality alignments. We also collect monolingual text for backtranslations. We follow tried and tested methods in neural machine translation using iterative backtranslation \cite{edunov-etal-2018-understanding} and adapt the multilingual denoising autoencoder model mBART-25 \cite{liu-etal-2020-multilingual-denoising} for translation. 

\section{Datasets}
We used several parallel and monolingual datasets, both publicly available and created in-house.

\subsection{Parallel data}
The parallel data used are ParIce \cite{Steingrmsson2020EffectivelyAA} and the JW300 corpus 
\cite{agic-vulic-2019-jw300}. In addition we used a parallel student theses and dissertation abstracts corpus, IPAC, generated in-house and sourced from the Skemman repository\footnote{\url{https://skemman.is}} as described in \cite{simonarson2021icelandic}. A breakdown of the data is shown in Table \ref{tab:parcorp}.

\begin{table}[htb]
\begin{center}
\begin{tabular}{l r}
\hline \bf Corpus & \bf \#Sentences \\ 
\hline
The Bible &  33k \\
EEA regulatory texts & 1,700k \\
EMA & 404k \\
ESO & 12.6k \\
OpenSubtitles &  1,300k \\
Tatoeba & 10k \\
Jehova's Witnesses (JW300) & 527k \\
Other$^\ast$ & 93k \\
IPAC & 64k \\
\hline
\end{tabular}
\end{center}
\begin{center}
\caption{Parallel corpora used. \#Sentences are the number of sentence pairs. \emph{Other$^\ast$} denotes software localizations, Project Gutenberg literature and the Icelandic sagas.} \
\label{tab:parcorp}
\end{center}
\end{table}

Following \cite{Pinnis2018TildesPC} we apply simple heuristic filters to the parallel data, mainly for capturing OCR and PDF errors, and correcting or removing character encoding errors after deduplication. Filters include but are not limited to: empty sequence removal, length cut-offs, character whitelists, mismatch in case and symbols between languages, edit-distances between source and target, normalizing of punctuation, and ad-hoc regular expressions for Icelandic specific OCR/PDF errors. For a more in-depth description see \cite{Jnsson2020ExperimentingWD}.

Other potential parallel datasets are ParaCrawl \cite{banon-etal-2020-paracrawl} and CCMatrix \cite{schwenk-etal-2021-ccmatrix}. Manual review of a couple of hundred randomly chosen lines from ParaCrawl revealed that the data quality is quite low for Icelandic, many lines are machine translated or badly aligned. We therefore did not include ParaCrawl. CCMatrix did not exist when the project started and we have not taken the time to review and integrate it although a quick inspection indicates that the quality is higher than in ParaCrawl.



\subsection{Data used for backtranslation}
We collected and translated monolingual data for backtranslations, made available in \cite{backtrans_corpus}, mostly building on the work in \cite{edunov-etal-2018-understanding}. The English sentences (44.7m) are retrieved from the Wikipedia, Newscrawl and Europarl corpora. The Icelandic sentences (31.3m) are sourced from the Icelandic Gigaword Corpus \cite{steingrimsson_2018}.

\begin{table}[htb]
\centering
\begin{tabular}{l l r}
\hline
\textbf{Lang.} & \textbf{Name} & \textbf{\#Sentences} \\
\hline
IS & Court rulings & 1.8M \\
IS & Supreme court rulings & 2M \\
IS & Laws & 814k \\
IS & Web of Science & 268k \\
IS & Wikipedia & 405k \\
IS & Parliamentary proc. & 6.2M \\
IS & Misc & 350k \\
IS & Newspaper (Mbl) & 13.6M \\
IS & Newspaper (Visir) & 4.9M \\
IS & Radio transcripts & 1M \\ \hline
EN & Newscrawl & 33.4M \\
EN & Wikipedia & 9.3M \\
EN & Europarl & 2.0M \\
\end{tabular}
\caption{Monolingual data used for backtranslation.}
\label{tab:monol-bt}
\end{table}

\section{Training of small transformer models}
Our earlier models were trained using the transformer-base configuration described in \cite{vaswaniAttentionAllYou2017} as implemented in Google's Tensor2Tensor (TensorFlow-based) \cite{vaswani-etal-2018-tensor2tensor} library. For later models we switched to Facebook's Fairseq \cite{ott-etal-2019-fairseq} library. An improved translation task was implemented in Fairseq to include BPE dropout; it is available in the \texttt{greynirseq}\footnote{\url{https://github.com/mideind/greynirseq}} library.

The transformer-base models were trained iteratively and used to generate new backtranslations. We stopped when each language direction had been trained on backtranslations that were produced by a model that had itself seen backtranslations at training time. We compared tagged and untagged backtranslations, sampling versus beam search and different mixing ratios (upsampling rate) between parallel and backtranslated data. Using tagged backtranslations as opposed to no tag showed an improvement from 16.5 to 17.5 BLEU\footnote{SacreBLEU\,signature: \texttt{BLEU+case.mixed+\\lang.en-is+numrefs.1+smooth.exp+tok.13a+\\version.1.5.1}} after the first iteration over the IPAC development set, while using no backtranslations gave 15.0, so we proceeded to use tagged translations.







\begin{table}[htb]
\centering
\begin{tabular}{l r}
\hline
\textbf{Model} & \textbf{BLEU} \\
\hline
Transformer-base & 16.5    \\
Transformer-base + bt & 17.5   \\
Transformer-base + iterative-bt & 18.5  \\
mBART (first run) & 23.1 \\
mBART (continued) & 23.6 \\
\end{tabular}
\caption{BLEU scores over IPAC for the EN-IS direction.}
\label{tab:transformer-base}
\end{table}




We use the IPAC test set to measure BLEU since it was available, has a large range of topics (although maybe not a large range of style) and is very unlikely to be accidentally included in the training data. The IPAC data is out of distribution with the rest of the training data but we do not consider that to be a problem since our goal is a general purpose model. The WMT dev set did not exist at the time.

We used a joint BPE vocabulary of size 16k and shared input-output embedding matrices. We pre-tokenized the input using \texttt{tokenizer} \footnote{\url{https://github.com/mideind/Tokenizer}} for the Icelandic side and spaCy \cite{spacy} for the English side. A beam width of 4 was used for beam search during backtranslation. Each training iteration took approximately one week on a single GTX 1080 graphics card. We were pleasantly surprised with how far we got with only this modest hardware.



\subsection{Translation mixing ratio selection and beam noise}
We assessed the impact of the ratio of synthetic backtranslation data to authentic parallel data on translation performance. Best results were obtained with a 1:2 ratio of authentic to synthetic data, using IPAC (held out from training) for evaluation.

For noising the backtranslation beam outputs, we follow \cite{edunov-etal-2018-understanding} and used within-$k$ permutation of whole words (with $k$=3), whole-word masking, and word dropout. Using sampling and noised beam outputs yielded comparable results. 


%

\section{Adapting mBART-25 for translation}
The mBART-25 \cite{liu-etal-2020-multilingual-denoising} (610M parameters) language model is far larger than the Transformer-base model (110M parameters). It was pretrained on 25 languages, including English and Swedish, but not Icelandic. We adapt it for translation from Icelandic to English and vice versa, using the same human-derived parallel translation data as for the transformer-base model along with the synthetic backtranslated corpus in a ratio of 1:2. We do not use any pre-tokenization and inherit the BPE sententencepiece vocabulary from mBART-25 (of size 250k) with the addition of an Icelandic language marker that was randomly initialized. We use the same hyperparameters as in \cite{liu-etal-2020-multilingual-denoising} and the implementation from Fairseq \cite{ott-etal-2019-fairseq}. The models are trained until their performance on the development sets plateaus.

The initial learning rate was set to 3e-4. Sixteen 32GB nVidia V100 GPUs connected with Infiniband were used for training. The effective batch size was around 10k sequences and the training took around 4 days of wall clock time per model.

Subsequently, these trained models were used to generate improved backtranslations. We then continued training the first iteration of our models with the new backtranslated data for another 30,000 steps for the Icelandic-English direction, and 36,000 steps for the English-Icelandic direction. The same training configurations were maintained as for the earlier runs.

\begin{table}[htb]
\centering
\begin{tabular}{l r r r r}
\hline
\textbf{Dir.} & \textbf{Steps} & \textbf{'21 test} & \textbf{'21 dev} & \textbf{EEA} \\
\hline
En-Is & 40k & 22.7 & 25.9 & 54.5  \\
En-Is & 40k + 36k & 24.3 & 27.8 & 57.6 \\ \hline
Is-En & 36k & 32.9 & 30.4 & 61.0 \\
Is-En & 36k + 30k & 33.5 & 31.8 & 63.2 \\
\end{tabular}

\caption{BLEU scores for the mBART-25 adapted translation models over the newstest2021 and EEA evaluation sets.}
\label{tab:mbart}
\end{table}

The benefit of continuing training of the mBART-derived models ranges from 0.6 to 3.1 BLEU as shown in Table \ref{tab:mbart}. BLEU performance is shown for both the newstest2021 development set as well as our cleaned-up dataset with sentence pairs from the EEA regulation corpus. Note that we do not finetune prior to evaluation nor do we perform checkpoint averaging.


\section{Conclusion}
We have shown how a small team with modest resources can adapt state-of-the-art methods to a medium resource language and achieve competitive results on machine translation between English and Icelandic.

The trained models are available for translation at \url{https://velthyding.is} and will be made available at the open \texttt{CLARIN-IS}\footnote{\url{https://repository.clarin.is}} repository. While a formal human comparison of the current models to the popular Google Translate service has not been performed, hundreds of monthly active users choose our solutions for translation between Icelandic and English.

\section{Future work}
We note the relatively small training time of the mBART adaptation and the lack of Icelandic data in the pretraining task for mBART as primary factors that can be addressed for improving results. Additionally online (or semi-online) self-training instead of train-then-translate would also improve results, especially with selective loss truncation as described in \cite{zhou-etal-2021-detecting}. The data selected for backtranslation should also be expanded for greater diversity of both genre and vocabulary. Finally, extending the translation context beyond the current sentence level is likely to improve results.


\section*{Acknowledgements}
We would like to thank Prof. Dr.-Ing. Morris Riedel and his team for providing access to the supercomputer at Forschungszentrum Jülich where the mBART-25 translation models were trained.

This project was supported by the Language Technology Programme for Icelandic 2019–2023. The programme, which is managed and coordinated by Almannarómur, is funded by the Icelandic Ministry of Education, Science and Culture.

\bibliography{anthology,custom}
\bibliographystyle{acl_natbib}

\end{document}